\documentclass[conference]{IEEEtran}
\IEEEoverridecommandlockouts
\usepackage{times}
\usepackage{geometry}
\usepackage[numbers]{natbib}
\usepackage{multicol}
\usepackage[bookmarks=true]{hyperref}

\usepackage{url}
\usepackage{times}
\usepackage{algorithmic}
\usepackage{amsmath}
\usepackage{amssymb}
\usepackage{amsfonts}
\usepackage{amsthm}
\usepackage{mathrsfs}
\usepackage{subfigure}
\usepackage{wrapfig}
\usepackage{comment}
\usepackage{xspace}
\usepackage{enumerate}
\usepackage{epsfig}
\usepackage{epstopdf}
\usepackage{booktabs}
\usepackage{multicol}
\usepackage{graphicx}
\usepackage{caption}
\usepackage{comment}
\usepackage{tabularx}
\usepackage[ruled,linesnumbered]{algorithm2e}
\usepackage{algorithmic}
\usepackage[nolist]{acronym}
\usepackage[dvipsnames]{xcolor}

\DeclareMathOperator{\sig}{sig}
\DeclareMathOperator{\tr}{\rm{tr}}

\begin{acronym}
\acro{DNN}{Deep Neural Network}
\acro{DP}{Dynamic Programming}
\acro{FBSDE}{Forward-Backward Stochastic Differential Equation}
\acro{LSTM}{Long-Short Term Memory}
\acro{FC}{Fully Connected}
\acro{DDP}{Differential Dynamic Programming}
\acro{HJB}{Hamilton-Jacobi-Bellman}
\acro{PDE}{Partial Differential Equation}
\acro{PI}{Path Integral}
\acro{NN}{Neural Network}
\acro{GPs}{Gaussian Processes}
\acro{SOC}{Stochastic Optimal Control}
\acro{RL}{Reinforcement Learning}
\acro{MPOC}{Model Predictive Optimal Control}
\acro{IL}{Imitation Learning}
\acro{RNN}{Recurrent Neural Network}
\acro{DL}{Deep Learning}
\acro{SGD}{Stochastic Gradient Descent}
\end{acronym}

\begin{document}

\title{Learning Deep Stochastic Optimal Control Policies using Forward-Backward SDEs}


\author{\authorblockN{Marcus A. Pereira$^{1\dagger*}$, Ziyi Wang$^{2*}$, Ioannis Exarchos$^{3}$ and Evangelos A. Theodorou$^{1,2}$}
\thanks{$^{1}$Institute for Robotics and Intelligent Machines, Georgia Institute of Technology}\thanks{$^{2}$The Center for Machine Learning, Georgia Institute of Technology}\thanks{$^{3}$School of Medicine, Emory University}\thanks{$^{*}$Equal contribution}\thanks{$^\dagger$Correspondence to Marcus A. Pereira: \href{mailto:mpereira30@gatech.edu}{mpereira30@gatech.edu} }}


%

\maketitle

\begin{abstract}
In this paper we propose a new methodology for decision-making under uncertainty using recent advancements in the areas of nonlinear stochastic optimal control theory,  applied mathematics, and machine learning. Grounded on the fundamental relation between certain nonlinear partial differential equations and forward-backward stochastic differential equations,  we develop a control framework that is scalable and applicable to general classes of stochastic systems and decision-making problem formulations in robotics and autonomy. The proposed deep neural network architectures for stochastic control consist of recurrent and fully connected layers. The performance and scalability of the aforementioned algorithm are investigated in three non-linear systems in simulation with and without control constraints. We conclude with a discussion on future directions and their implications to robotics. 
\end{abstract}

\IEEEpeerreviewmaketitle

\section{Introduction}

  Over the past 15 years there has been significant interest from the robotics community in developing algorithms for stochastic control of systems operating in dynamic and uncertain environments. This interest was initiated  by two main developments related to theory and  hardware. From a theoretical standpoint,  there has been a better -- and in some sense  deeper --  understanding of  connections between different disciplines. As an example, the connections between optimality principles in control theory and information theoretic concepts in statistical physics  are well understood so far \cite{DaiPra1996,Fleming1971,TheodorouCDC2012,entropy_2015}. These connections  have resulted in novel  algorithms that are scalable, real-time, and can handle complex nonlinear dynamics \cite{TRO_Grady_2018}.  On the hardware side, there have been significant technological  developments   that  made possible the use of    high performance computing for real-time \ac{SOC} in robotics \cite{NVIDIA1999}.

Traditionally,  \ac{SOC} problems are solved using \ac{DP}. \acl{DP}  requires solving a nonlinear second order \ac{PDE} known as the \ac{HJB} equation \cite{bellman2013dynamic}. It is well-known that the \ac{HJB} equation suffers from the curse of dimensionality. One way to tackle this problem is through an exponential transformation to linearize the \ac{HJB} equation, which can then be solved with forward sampling using the linear Feynman-Kac lemma \cite{TheodorouGPI} \cite{Karatzasbook}. While the linear Feynman-Kac lemma provides a probabilistic representation of the solution to the \ac{HJB} that is exact, its application relies on  certain assumptions between control authority and noise. In addition, the exponential transformation of the value function reduces the discriminability between \textit{good} and \textit{bad} states, which makes the computation of the optimal control policy difficult.

An  alternative  approach to solve \ac{SOC} problems is to transform the HJB into  a system of \acp{FBSDE} using a nonlinear version of the Feynman-Kac lemma \cite{yong1999stochastic,Pardoux_Book2014}. This is a more general approach compared to the standard Path Integral control framework, in that it does not rely on any assumptions between control authority and noise. In addition, it is valid for general classes of stochastic processes  including jump-diffusions and  infinite dimensional stochastic processes \cite{kharroubi2015,fabbri}. However, the main challenge in using the nonlinear Feynman-Kac lemma lies in the solution of the backward SDE. This process requires the back-propagation of a conditional expectation, and thus cannot be solved by simple trajectory integration, as it is done with forward SDEs. Therefore, numerical approximation techniques are needed for utilization in an actual algorithm. \citet{exarchos2018stochastic} developed an importance sampling based iterative scheme by approximating the conditional expectation at every time step using linear regression (see also \cite{exarchos2016learning} and \cite{exarchos2017stochastic}). However, this method suffers from compounding errors from Least Squares approximation at every time step.

Recently, the idea of using \acp{DNN} and other data-driven techniques for approximating the solutions of non-linear \acp{PDE} has been garnering significant attention. In  \citet{raissi2019physics}, \acp{DNN} were used for both solving and data-driven discovery of the coefficients of non-linear \acp{PDE} popular in physics literature such as the Schr\"odinger, the Allen-Cahn, the Navier-Stokes, and the Burgers equations. They have demonstrated that their \ac{DNN}-based approach can surpass the performance of other data-driven methods such as sparse linear regression proposed by \citet{rudy2017data}. On the other hand, using \acp{DNN} for end-to-end \ac{MPOC} has also become a popular research area. \citet{pereira2018mpc} introduced a \ac{DNN} architecture for \ac{IL}, inspired by \ac{MPOC}, based on the \ac{PI} Control approach alongside \citet{amos2018differentiable} who introduced an end-to-end \ac{MPOC} architecture that uses the KKT conditions of the convex approximation. \citet{pan2017agile} demonstrated the \ac{MPOC} capabilities of a \ac{DNN} control policy using only camera and wheel speed sensors, through \ac{IL}. \citet{morton2018deep} used a Koopman operator based \ac{DNN} model for learning the dynamics of fluids and performing \ac{MPOC} for suppressing vortex shedding in the wake of a cylinder.  

This tremendous success of \acp{DNN} as universal function approximators \cite{GoodfellowDL} inspires an alternative scheme to solve systems of \acp{FBSDE}. Recently, \citet{Han8505} introduced a Deep Learning based algorithm to solve \acp{FBSDE} associated with nonlinear parabolic \acp{PDE}. Their framework was applied to solve the \ac{HJB} equation for a white-noise driven linear system to obtain the value function at the initial time step. This framework, although effective for solving parabolic PDEs, can not be applied directly to solve the HJB for optimal control of unstable nonlinear systems since it lacks sufficient exploration and is limited to only states that can be reached by purely noise driven dynamics. This problem was addressed in \cite{exarchos2018stochastic} through application of Girsanov's theorem, which allows for the modification of the drift terms in the \ac{FBSDE} system thereby facilitating efficient exploration through controlled forward dynamics.

In this paper, we propose a  novel framework  for solving \ac{SOC} problems of  nonlinear systems in robotics. The resulting  algorithms  overcome limitations of previous work in \cite{Han8505} by exploiting Girsanov's theorem as in \cite{exarchos2018stochastic} to enable efficient exploration and by utilizing the benefits of recurrent neural networks in learning temporal dependencies. We begin by proposing essential modifications to the existing framework of \acp{FBSDE} to utilize the solutions of the \ac{HJB} equation at every timestep to compute an optimal feedback control which thereby drives the exploration to optimal areas of the state space. Additionally, we propose a novel architecture that utilizes \ac{LSTM} networks to capture the underlying temporal dependency of the problem. In contrast to the individual \ac{FC} networks in \cite{Han8505}, our proposed architecture uses fewer parameters, is faster to train, scales to longer time horizons and produces smoother control trajectories. We also extend our framework to problems with control-constraints which are very relevant to most applications in Robotics wherein actuation torques must not violate specified box constraints. Finally, we compare the performance of both network architectures on systems with nonlinear dynamics such as pendulum, cartpole and quadcopter in simulation.

The rest of this paper is organized as follows: in Section \ref{FBSDE Formulation} we reformulate the stochastic optimal control problem in the context of FBSDE. In Section \ref{Control constraint} we use the same FBSDE framework to the control constrained case. Then we provide the Deep \ac{FBSDE} Control algorithm in Section \ref{Algorithm}. The simulation results are included in Section \ref{experiments}. Finally we conclude the paper and discuss future research directions.

\section{Stochastic Optimal Control through \ac{FBSDE}}\label{FBSDE Formulation}
\subsection{Problem Formulation}
Let ($\Omega, \mathcal{F}, \{\mathcal{F}_t\}_{t\geq0},\mathbb{Q}$) be a complete, filtered probability space on which a $v$-dimensional standard Brownian motion $w(t)$ is defined, such that $\{\mathcal{F}_t\}_{t\geq0}$ is the normal filtration of $w(t)$. Consider a general stochastic nonlinear system with control affine dynamics,
\begin{equation}
    \mathrm{d}x(t) = f(x(t),t)\mathrm{d}t + G(x(t),t)u(x(t),t)\mathrm{d}t + \Sigma(x(t),t)\mathrm{d}w(t)
    \label{eqn:system_dyn}
\end{equation}
where, $0<t<T<\infty$, $T$ is the time horizon, $x\in\mathbb{R}^n$ is the state vector, $u\in\mathbb{R}^m$ is the control vector, $f:\mathbb{R}^n \times [0,T] \rightarrow \mathbb{R}^n$ represents the drift, $G:\mathbb{R}^n \times [0,T] \rightarrow\mathbb{R}^{n\times m}$ represents the actuator dynamics, $\Sigma:\mathbb{R}^n \times [0,T] \rightarrow \mathbb{R}^{n\times v}$ represents the diffusion. The \acl{SOC} problem can be formulated as minimization of an expected cost functional given by
\begin{equation}
    J\big(x(t),t\big) = \mathbb{E}_{\mathbb{Q}}\Big[g\big(x(T)\big) + \int_{t}^T \big(q(x(\tau)\big) + \frac{1}{2}u^\mathrm{T}Ru)\mathrm{d}\tau \Big],
    \label{eqn:cost}
\end{equation}
where $g: \mathbb{R}^n \rightarrow \mathbb{R}^+$ is the terminal state cost, $q: \mathbb{R}^n \rightarrow \mathbb{R}^+$ is the running state cost and $R$ is a $m\times m$ positive definite matrix. The expectation is taken with respect to the probability measure $\mathbb{Q}$ over the space of trajectories induced by controlled stochastic dynamics. With the set of all admissible controls $\mathcal{U}$, we can define the value function as,
\begin{equation}
    \begin{cases}
    V\big(x(t),t\big) &= \inf_{u(.) \in \mathcal{U}[0,T] }J\big(x(t),t\big) \\
    V\big(x(T), T\big) &= g\big(x(T)\big).
    \end{cases}
\end{equation}

Using stochastic Bellman's principle, as shown in \cite{yong1999stochastic}, if the value function is in $C^{1,2}$, then its solution can be found with Ito's differentiation rule to satisfy the \acl{HJB} equation,
\begin{equation}
    \begin{cases}
        V_t + \inf_{u(\cdot) \in \mathcal{U}[0,T]} \Big\{\frac{1}{2}\tr(V_{xx}\Sigma \Sigma^\mathrm{T})+V_x^\mathrm{T}(f+Gu)+q\\
        +\frac{1}{2}u^\mathrm{T}Ru\Big\}=0 \\
        V(x(T),T)=g(x(T)),
    \end{cases}
    \label{eqn:HJBoriginal}
\end{equation}
where $V_x, V_{xx}$ denote the gradient and Hessian of $V$ respectively. The explicit dependence on independent variables in the \ac{PDE} above and all \acp{PDE} henceforth is omitted for the sake of conciseness, but will be maintained for their corresponding SDEs for clarity. For the chosen form of the cost functional integrand, the infimum operation can be carried out by taking the gradient of the terms inside, known as the Hamiltonian, with respect to $u$ and setting it to zero,
\begin{equation}
    G^\mathrm{T}(x(t), t)V_x(x(t), t) + Ru(x(t),t) = 0.
\end{equation}
Therefore, the optimal control is obtained as
\begin{equation}
    u^*(x(t),t) = -R^{-1}G^\mathrm{T}(x(t),t)V_x(x(t),t).
    \label{eqn:optimalontrol}
\end{equation}
Plugging the optimal control back into the original \ac{HJB} equation, the following form of the equation is obtained,
\begin{equation}
    \begin{cases}
        V_t + \frac{1}{2}\tr(V_{xx}\Sigma\Sigma^\mathrm{T}) + V_x^\mathrm{T}f + q - \frac{1}{2}V_x^\mathrm{T}GR^{-1}G^\mathrm{T}V_x=0 \\
        V(x(T),T)=g(x(T)).
    \end{cases}
    \label{eqn:HJBfinal}
\end{equation}

\subsection{Non-linear Feynman-Kac lemma}
Here we restate the non-linear Feynman-Kac lemma from \cite{exarchos2018stochastic}. Consider the Cauchy problem,
\begin{equation}
    \begin{cases}
    \nu_t + \frac{1}{2} \tr\big(\nu_{xx}\Sigma\Sigma^\mathrm{T}\big) +  \nu_x^\mathrm{T} b + h = 0\\
    \nu(x(T),T) = g(x(T)),\, x \in \mathbb{R}^n,
    \end{cases}
    \label{eqn:parabolic_pde}
\end{equation}
wherein the functions $\Sigma(x(t),t)$, $b(x(t),t)$, $h(x,\nu, z,t)$ and $g(x(T))$ satisfy mild regularity conditions \cite{exarchos2018stochastic}. Then, (\ref{eqn:parabolic_pde}) admits a unique (viscosity) solution $\nu: \mathbb{R}^n \times [0,T] \rightarrow \mathbb{R}$, which has the following probabilistic representation,
\begin{align}
    \nu(x(t),t)&=y(t)\\
    \Sigma^\mathrm{T}\nu_x(x(t),t)&=z(t)
\end{align}
wherein $\big(x(\cdot),y(\cdot),z(\cdot) \big)$ is the unique solution of an \ac{FBSDE} system. The forward component of that system is given by 
\begin{equation}
\begin{cases}
    \mathrm{d}x(t) &= b(x(t),t)\mathrm{d}t + \Sigma(x(t),t)\mathrm{d}w(t)
    \\ x(0)&=\xi \\
\end{cases}
\label{eqn:FSDE}
\end{equation}
where, without loss of generality, $w$ is chosen as a n-dimensional Brownian motion. The process $x(t)$, satisfying the above forward SDE, is also called the \textit{state process}. The associated backward SDE is
\begin{equation}
\begin{cases}
    \mathrm{d}y(t) &= - h(x(t),y(t),z(t),t)\mathrm{d}t + z(t)^\mathrm{T} \mathrm{d}w(t)\\
    y(T) &= g(x(T)). 
\end{cases}
\label{eqn:BSDE}
\end{equation}
The function $h(\cdot)$ is called the \textit{generator} or \textit{driver}.

We assume that there exists a matrix-valued function $\Gamma:\mathbb{R}^n\times[0,T]\rightarrow \mathbb{R}^{n\times m}$ such that the controls matrix $G(x(t),t)$ in (\ref{eqn:system_dyn}) can be decomposed as $G(x(t),t)=\Sigma(x(t),t)\Gamma(x(t),t)$ for all $(x,t)\in\mathbb{R}^n\times [0,T]$, satisfying the same mild regularity conditions. This decomposition can be justified as the case of stochastic actuators, where noise enters the system through the control channels. Under this assumption, we can apply the nonlinear Feynman-Kac lemma to the \ac{HJB} \ac{PDE} (\ref{eqn:HJBfinal}) and establish equivalence to (\ref{eqn:parabolic_pde}) with coefficients of (\ref{eqn:parabolic_pde}) given by 
\begin{equation}\label{eqn:valueFnHJBEquivalence}
\begin{split}
    b(x(t),t) &= f(x(t),t)\\
    h(x(t),y(t),z(t),t) &= q(x(t)) - \frac{1}{2} z^\mathrm{T} \Gamma R^{-1} \Gamma^\mathrm{T} z.
\end{split}
\end{equation}

\subsection{Importance Sampling for Efficient Exploration}
There are several cases of systems in which the goal state practically cannot be reached by the uncontrolled stochastic system dynamics. This issue can be eliminated if one is given the ability to modify the drift term of the forward SDE. Specifically, by changing the drift, we can direct the exploration of the state space towards the given goal state, or any other state of interest, reachable by control. Through Girsanov's theorem \cite{Girsanov} on change of measure, the drift term in the forward SDE (\ref{eqn:FSDE}) can be changed if the backward SDE (\ref{eqn:BSDE}) is compensated accordingly. This is known as the importance sampling for \acp{FBSDE}. This results in a new system of \acp{FBSDE} in certain sense equivalent to the original ones,
\begin{equation}
    \begin{cases}
    \mathrm{d}\tilde{x}(t) = [b(\tilde{x}(t),t) + \Sigma(\tilde{x}(t),t)K(t)]\mathrm{d}t + \Sigma(\tilde{x}(t),t)\mathrm{d}\tilde{w}(t)\\
    \tilde{x}(0) = \xi,
    \end{cases}
    \label{eqn:new_fsde}
\end{equation}
along with the compensated BSDE,
\begin{equation}
    \begin{cases}
    \mathrm{d}\tilde{y}(t) = (-h(\tilde{x}(t), \tilde{y}(t), \tilde{z}(t), t) + \tilde{z}(t)^\mathrm{T}K(t))\mathrm{d}t \\+ \tilde{z}(t)^\mathrm{T}\mathrm{d}\tilde{w}(t)\\
    \tilde{y}(T) = g(\tilde{x}(T)),
    \end{cases}
    \label{eqn:new_bsde}
\end{equation}
for any measurable, bounded and adapted process $K:[0, T] \rightarrow \mathbb{R}^n$. We refer the readers to proof of Theorem 1 in \cite{exarchos2018stochastic} for the full derivation of change of measure for \acp{FBSDE}. The PDE associated with this new system is given by
\begin{equation}
    \begin{cases}
        V_t + \frac{1}{2}\tr\big(V_{\tilde{x}\tilde{x}}\,\Sigma \Sigma^\mathrm{T}\big) +  V_x^\mathrm{T} \big(b +\Sigma K\big) + h - \tilde{z}^\mathrm{T} K = 0\\
        V(\tilde{x},T) = g(\tilde{x}(T)),
    \end{cases}
    \label{eqn:new_pde}
\end{equation}
which is identical to the original problem (\ref{eqn:parabolic_pde}) as we have merely added and subtracted the term $\tilde{z}^\mathrm{T} K$. Recalling the decomposition of control matrix in the case of stochastic actuators, the modified drift term can be applied with any nominal control $\bar{u}$ to achieve the controlled dynamics,
\begin{equation}
\begin{split}
    \mathrm{d}\tilde{x}(t) = \big[f(\tilde{x}(t), t) + \Sigma\big(\tilde{x}(t), t\big)&\Gamma\big(\tilde{x}(t), t\big)\bar{u}(t)\big]\mathrm{d}t \\
    &+ \Sigma\big(\tilde{x}(t), t\big)\mathrm{d}\tilde{w}(t)
\end{split}
\end{equation}
with $K(t)=\Gamma(\tilde{x}(t),t)\, \bar{u}$. 
The nominal control $\bar{u}$ can be any open or closed-loop control, a random control, or a control calculated from a previous run of the algorithm.

\subsection{FBSDE Reformulation}
Solutions to BSDEs need to satisfy a terminal condition, and thus, integration needs to be performed backwards in time, yet the filtration still evolves forward in time. It turns out that a terminal value problem involving BSDEs admits an adapted solution if one back-propagates the conditional expectation of the process. This was the basis of the approximation scheme and corresponding algorithm introduced in \cite{exarchos2018stochastic}. However, this scheme is prone to approximation errors introduced by least squares estimates which compound over time steps. 
On the other hand, the \ac{DL}-based approach in \cite{Han8505} uses the terminal condition of the BSDE as a prediction target for a self-supervised learning problem with the goal of using back-propagation to estimate the value function at the initial timestep. This was achieved by treating the value at the initial timestep, $V(\tilde{x}(0),0)$, as one of the trainable parameters of a \ac{DL} model. There is a two-fold advantage of this approach: (i) starting with a random guess of $V(\tilde{x}(0),0; \phi)$, the backward SDE can be forward propagated instead. This eliminates the need to back-propagate a least-squares estimate of the conditional expectation to solve the BSDE and instead treat the BSDE similar to the FSDE, and (ii) the approximation errors at every time step are compensated by the backpropagation training process of DL. This is because the individual networks, at every timestep, contribute to a common goal of predicting the target terminal condition and are jointly trained.  

\begin{figure*}
\centering
  \includegraphics[width=0.8\linewidth]{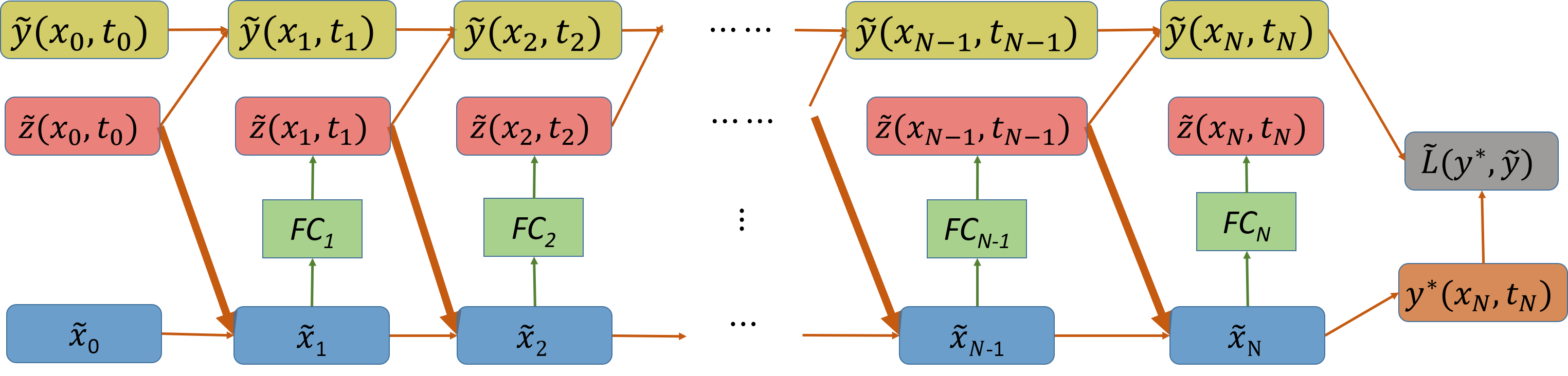}
  \caption{\textbf{\ac{FC} neural network architecture} (boldfaced connections indicate importance sampling).}
\label{fig:FCnetwork}
\end{figure*}

\begin{figure*}
\centering
  \includegraphics[width=0.8\linewidth]{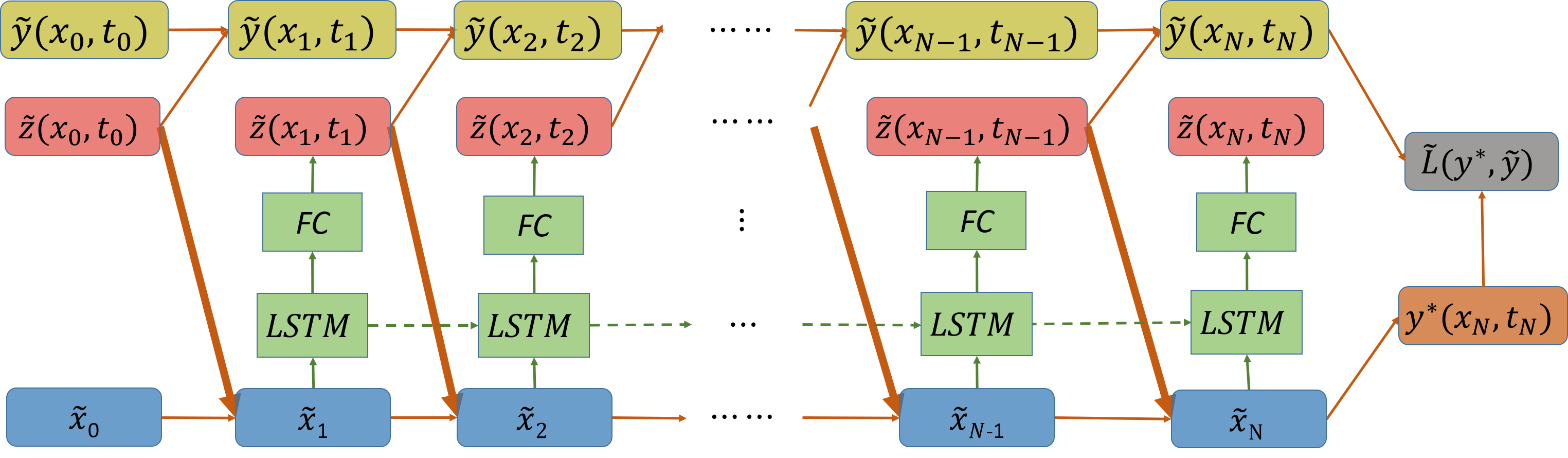}
  \caption{\textbf{\ac{LSTM} neural network architecture} (boldfaced connections indicate importance sampling). Note that the $FC_{t}$ networks in (Fig. \ref{fig:FCnetwork}) above are different for each time step, whereas here the same weights are shared for every time step.}
\label{fig:LSTMnetwork}
\end{figure*}

In this work, we combine the importance sampling concepts for \acp{FBSDE} with the \acl{DL} techniques that allows for the forward sampling of the BSDE and propose a new algorithm for \acl{SOC} problems. The novelty of our approach is to incorporate importance sampling for efficient exploration in the \ac{DL} model. Instead of the original \ac{HJB} equation (\ref{eqn:HJBfinal}), we focus on obtaining solutions for the modified \ac{HJB} \ac{PDE} in (\ref{eqn:new_pde}) by using the modified \ac{FBSDE} system (\ref{eqn:new_fsde}), (\ref{eqn:new_bsde}). Additionally, we explicitly compute the control at every time step using the analytical expression for optimal control (\ref{eqn:optimalontrol}) in the computational graph. Similar to \cite{Han8505}, the \ac{FBSDE} system is solved by integration of both the SDEs forward in time as follows, 

\begin{equation}
    \begin{cases}
    \mathrm{d}\tilde{x}(t) = f(\tilde{x}(t),t)\mathrm{d}t + \Sigma(\tilde{x}(t),t) \big[ \bar{u}(\tilde{x}(t),t)\mathrm{d}t + \mathrm{d}\tilde{w}(t)\big]\\
    \bar{u}(\tilde{x}(t),t) = \Gamma u^* (\tilde{x}(t),t;\theta_t) = - \Gamma R^{-1} \Gamma^\mathrm{T} \tilde{z}(t;\theta_t)\\
    \tilde{x}(0) = \xi
    \end{cases}
    \label{eqn:newfsdecontrol}
\end{equation}
and
\begin{equation}
    \begin{cases}
    \mathrm{d}\tilde{y}(t) = \big(-h(\tilde{x}(t), \tilde{y}(t), \tilde{z}(t;\theta_t), t) \\+\tilde{z}(t;\theta_t)^\mathrm{T}\,\Gamma(\tilde{x}(t),t)\,\bar{u}\big)\mathrm{d}t + \tilde{z}(t;\theta_t)^\mathrm{T}\mathrm{d}\tilde{w}(t)\\
    \tilde{y}(0) = V(\tilde{x}(0),0;\phi).
    \end{cases}
    \label{eqn:newbsdecontrol}
\end{equation}

\section{Stochastic Control Problems with Control Constraints}\label{Control constraint}
  
The framework we have considered so far can be suitably modified to accommodate a certain type of control constraints, namely upper and lower bounds $(-u^{\max},u^{\max})$. Specifically, each control dimension component satisfies $|u_j(\tilde{x}(t),t)|\leq u^{\max}_j$ for all $j=\{1,\cdots,m\}$. Such control constraints are common in mechanical systems, where control forces and/or torques are bounded, and may be readily introduced in our framework via the addition of a ``soft'' constraint, integrated within the cost functional. In recent work, \citet{exarchosL1} showed how box-type control constraints for $L^1$-optimal control problems (also called \textit{minimum fuel} problems), can be incorporated into an \ac{FBSDE} scheme. These are in contrast to the more frequently used quadratic control cost ($L^2$ or \textit{minimum energy}) \ac{SOC} problems. Indeed, one can replace the cost functional given by (\ref{eqn:cost}) with
.\begin{equation}\label{cost2}
J\big(\tilde{x}(t), t\big)=\mathbb{E}_{\mathbb{Q}} \bigg[ g(\tilde{x}(T)) +  \int_{t}^{T} \bigg(q(\tilde{x}(t))+ \sum_{j=1}^m S_j(u_j) \bigg)  \mathrm{d}  t \bigg],
\end{equation} 
where 
\begin{align}
	S_j(u_j)=c_j\int_{0}^{u_j}\sig^{-1}\big(\frac{\mathrm{v}}{u^{\max}_j}\big) \mathrm{d} \mathrm{v}, \qquad j=\{1,\ldots,m\},
\end{align}
$c_j$ are constant weights, $\sig(\cdot)$ denotes the sigmoid (tanh-like) function that saturates at infinity, i.e., $\sig(\pm\infty)=\pm1$, while $\mathrm{v}$ is a dummy variable of integration. A suitable example along with its inverse is
\begin{align}
&	\sig(\mathrm{v})=\frac{2}{1+e^{-\mathrm{v}}}-1, \quad \mathrm{v}\in \mathbb{R} \\
&	\sig^{-1}(\mathrm{\mu})=\log\bigg(\frac{1+\mathrm{\mu}}{1-\mathrm{\mu}}\bigg),\quad \mathrm{\mu}\in (-1,1).
\end{align}

Following the same procedure as in Section \ref{FBSDE Formulation}, we set the derivative of the Hamiltonian equal to zero and obtain
\begin{equation}
	-\begin{bmatrix}
	c_1 \sig^{-1}(\frac{u_1}{u^{\max}_1})\\
	\vdots\\
	c_m \sig^{-1}(\frac{u_m}{u^{\max}_m})
	\end{bmatrix}-G^{\mathrm{T}}(\tilde{x}(t),t)v_{\tilde{x}}(\tilde{x}(t),t)=0.
\end{equation}

By introducing the notation 
$$G(\tilde{x}(t),t)=[\mathrm{g}_1(\tilde{x}(t),t)\quad \mathrm{g}_2(\tilde{x}(t),t)\quad \cdots \quad \mathrm{g}_m(\tilde{x}(t),t)]$$ where $\mathrm{g}_i$ (not to be confused with the terminal cost $g$) denotes the i-th column of $G$,
we may write the optimal control in component-wise notation as
\begin{align}\begin{split}
& u_j^*(\tilde{x}(t),t)=u^{\max}_j\sig\bigg(-\frac{1}{c_j}\mathrm{g}_j^{\mathrm{T}} (\tilde{x}(t),t)V_{\tilde{x}}(\tilde{x}(t),t)\bigg),\\
& j=\{1,\cdots,m\}
\end{split}
\end{align}

The optimal control can be written equivalently in vector form. Indeed, if $[u^{\max}_1,\ldots,u^{\max}_m]^\mathrm{T}$ is the vector of bounds, $R^{-1}=[1/c_1,\ldots,1/c_m]$ is a diagonal matrix of the reciprocals of the weights and $\mathrm{U}_{max}=diag([u^{\max}_1,\ldots,u^{\max}_m]^\mathrm{T})$ is a diagonal matrix of the bounds, one readily obtains
\begin{align}\label{eqn:optimalcontrolconstrained}
\begin{split}
	&u^*(\tilde{x}(t),t)=\mathrm{U}_{max}\sig\bigg(-R^{-1}G^{^\mathrm{T} }(\tilde{x}(t),t)V_{\tilde{x}}(\tilde{x}(t),t)\bigg)
\end{split}
\end{align}

Substituting the equation of the constrained controls into eqn. \ref{eqn:new_pde} results in
\begin{equation}
    \begin{cases}
        V_t + \frac{1}{2}\tr\big(V_{\tilde{x}\tilde{x}}\,\Sigma \Sigma^\mathrm{T}\big) +  V_{\tilde{x}}^\mathrm{T} \big(b +\Sigma K\big) + h - \tilde{z}^\mathrm{T} K = 0\\
        V(\tilde{x}(T),T) = g(\tilde{x}(T))
    \end{cases}
\end{equation}
where $ h $ is specified by the expression that follows: 
\begin{align}\label{eqn:valueFnHJBEquivalenceCC}
    h = q(\tilde{x}(t)) + V_{\tilde{x}}^{\mathrm{T}} G(\tilde{x}(t),t) u^*(\tilde{x}(t),t) +   \sum_{j=1}^m S_j(u_j^{*})
\end{align}

\begin{algorithm}
\caption{Finite Horizon Deep FBSDE Controller}
    \begin{algorithmic}
          \STATE \textbf{Given:}\\ 
          $\tilde{x}_0=\xi,\;f,\; G$, $\Sigma,\; \Gamma $: Initial state and system dynamics;\\ $g,\; q, \; R$: Cost function parameters;\\$N$: Task horizon, $K$: Number of iterations, $M$: Batch \\size; bool: Boolean for constrained control case;\\$U_{max}$: maximum controls per input channel; \\ $\Delta t$: Time discretization;
          $\lambda$: weight-decay parameter;
          \STATE \textbf{Parameters:}\\
          $\tilde{y}_0=V(\tilde{x}_0,0;\phi)$: Value function at $t=0$;\\
          $\tilde{z}_0=\Sigma^\mathrm{T} \,\nabla_{\tilde{x}} V $: Gradient of value function at $t=0$;\\
          $\theta$: Weights and biases of all fully-connected and/or \\LSTM layers;
          \STATE \textbf{Initialize neural network parameters};
          \STATE \textbf{Initialize states:}\\ $\{\tilde{x}^i_0\}_{i=1}^{M},\;\tilde{x}^i_0=\xi $\\
          $\{\tilde{y}^i_0\}_{i=1}^{M},\;\tilde{y}^i_0=V(\tilde{x}^i_0,0;\phi)$\\
          $\{\tilde{z}^i _0\}_{i=1}^{M},\;\tilde{z}^i _0=\Sigma^\mathrm{T} \nabla_{\tilde{x}} V(\tilde{x}^i_0,0;\phi)$
          \FOR{$k=1$ \TO $K$}
            \FOR{$i=1$ \TO $M$}
                \FOR{$t=1$ \TO $N-1$}
                    \STATE Compute gamma matrix: $\Gamma_t^i=\Gamma \big(\tilde{x}^i_t, t\big)$;   
                    \IF{bool == True}
                    \STATE ${u^i_t}^*=U_{max}sig\big(-R^{-1} {\Gamma^i_t}^\mathrm{T}\tilde{z}^i_t\big)$;
                    \ELSE
                    \STATE ${u^i_t}^*=-R^{-1}{\Gamma^i_t}^\mathrm{T} \tilde{z}^i_t$;
                    \ENDIF
                    \STATE Sample Brownian noise: $\Delta \tilde{w}^i_t \sim \mathcal{N}(0, \Sigma)$
                    \STATE Update value function: $\tilde{y}^i_{t+1} = \tilde{y}^i_t - \tilde{h}\big(\tilde{x}^i_t, \, \tilde{y}^i_t,\,\tilde{z}^i_t,t \big) \Delta t + {\tilde{z}^i_t}{}^{\mathrm{T}} \, \Gamma^i_t \, {u^i_t}{}^* \, \Delta t + {\tilde{z}^i_t}{}^{\mathrm{T}} \Delta \tilde{w}^i_t$
                    \STATE Update system state: $\tilde{x}^i_{t+1}=\tilde{x}^i_t + f(\tilde{x}^i_t,t) \Delta t + \Sigma \big(\Gamma^i_t {u^i_t}{}^*\Delta t +\Delta \tilde{w}^i_t\big) $
                    \STATE Predict gradient of value function: $\tilde{z}^i_{t+1} = f_{FC}\big(\tilde{x}^i_{t+1}; \theta_t^k\big)$ or $f_{LSTM}\big(\tilde{x}^i_{t+1}; \theta^k\big)$
                \ENDFOR
                \STATE Compute target terminal value: ${y^*_N}^i=g\big(\tilde{x}^i_N\big)$
            \ENDFOR
            \STATE Compute mini-batch loss: $\mathcal{L}= \displaystyle \frac{1}{M} \sum_{i=1}^M \| {y^*_N}^i - \tilde{y}^i_N\|^2_2 + \lambda \| \theta^k \|^2_2 $
            \STATE $\theta^{k+1}\leftarrow$ Adam.step($\mathcal{L}, \theta^k$); $\phi^{k+1}\leftarrow$ Adam.step($\mathcal{L},\phi^k$)           
          \ENDFOR
          \RETURN $\theta^K, \phi^K$
    \end{algorithmic}
    \label{alg:FBSDEalgorithm}
\end{algorithm}
 
\section{Deep FBSDE Controller}\label{Algorithm}
In this section we present the algorithm for the Deep FBSDE stochastic controller and discuss the underlying network architectures.  

\textbf{Algorithm:} The task horizon $0<t<T$ in continuous-time can be discretized as $t=\{0,1,\cdots,N\}$, where $T=N\Delta t$. Here we abuse the notation $t$ as both the continuous time variable and discrete time index. With this we can also discretize all the variables as step functions such that $\tilde{x}_t,\tilde{y}_t,\tilde{z}_t,u^*_t=\tilde{x}(t),\tilde{y}(t),\tilde{z}(t),u^*(t)$ if the discrete time index $t$ is between the time interval $\big[t\Delta t,(t+1)\Delta t\big)$.

\begin{figure*}
\centering
  \includegraphics[width=0.8\linewidth]{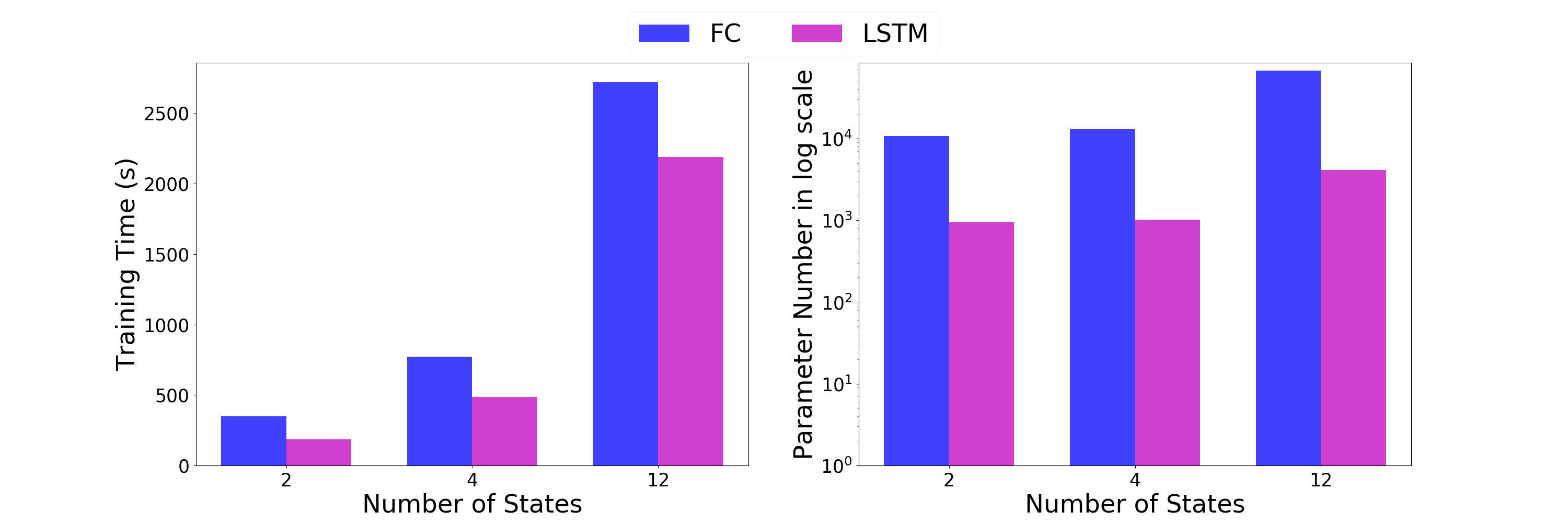}
  \caption{Comparing neural network training time \textit{(left)} and number of trainable parameters \textit{(right)} vs. state dimensionality  for the proposed FC (Fig. \ref{fig:FCnetwork}) and LSTM (Fig. \ref{fig:LSTMnetwork}) network architectures.}
\label{fig:run_time}
\end{figure*}

The Deep \ac{FBSDE} algorithm, as shown in Alg. \ref{alg:FBSDEalgorithm}, solves the finite time horizon control problem by approximating the gradient of the value function $\tilde{z}^i_t$ at every time step with a \ac{DNN} parameterized by $\theta_t$. Note that the superscript $i$ is the batch index, and the batch-wise calculation can be implemented in parallel. The initial value $\tilde{y}^i_0$ and its gradient $\tilde{z}^i_0$ are parameterized by trainable variables $\phi$ and are randomly initialized. The optimal control action is calculated using the discretized version of (\ref{eqn:optimalontrol}) (or (\ref{eqn:optimalcontrolconstrained}) for the control constrained case).
The dynamics $\tilde{x}$ and value function $\tilde{y}$ are propagated using the Euler integration scheme, as shown in the algorithm. The function $h$ is calculated using (\ref{eqn:valueFnHJBEquivalence}) (or (\ref{eqn:valueFnHJBEquivalenceCC}) for the control constrained case). The predicted final value $\tilde{y}^i_N$ is compared against the true final value ${y^*_N}^i$ to calculate the loss. The networks can be trained with any one of the variants of \ac{SGD} such as the Adam optimizer \cite{Adam} until convergence with custom learning rate scheduling. The trained networks can then be used to predict the optimal control at every time step starting from the given initial condition $\xi$.

\textbf{Network Architectures:} The network architectures illustrated in figures \ref{fig:FCnetwork} and \ref{fig:LSTMnetwork}, are extensions of the network introduced in \cite{Han8505} (refer to fig. 4 in the paper). The neural network architectures in figures \ref{fig:FCnetwork} and \ref{fig:LSTMnetwork} have additional connections (highlighted by boldfaced arrows)  that use the predicted gradient of the value function at every time step to compute and apply an optimal feedback control. An architecture similar to fig. \ref{fig:FCnetwork} was introduced in \cite{han2016deep} to solve model-based \ac{RL} problems posed as finite time horizon \ac{SOC} problems. This consisted of a \ac{FC} network at every timestep to predict an action as a function of the current state. The networks were stacked together to form one large deep network which was trained in an end-to-end fashion with the goal of minimizing the accumulated cost (or maximizing accumulated reward). In contrast, the network architecture in  fig. \ref{fig:FCnetwork}  uses the explicit form of the optimal feedback control (eq. \eqref{eqn:optimalontrol} or eq. \eqref{eqn:optimalcontrolconstrained}) at every timestep calculated using the value function gradient predicted by the network. In addition, we use the prediction to propagate the value function according to the BSDE \eqref{eqn:newbsdecontrol} and minimize the difference between the propagated value function and the true value function at the final state. This, however, creates a new path for gradient backpropagation through time \cite{werbos1990backpropagation} which introduces both advantages and challenges for training the networks. The advantage being a direct influence of the weights on the state cost $q(\tilde{x}_t)$ leading to accelerated convergence. Nonetheless, this passage also leads to the vanishing gradient problem, which has been known to plague training of \acp{RNN} for long sequences (or time horizons). 

To tackle this problem, we propose a new LSTM-based network architecture, as shown in fig. \ref{fig:LSTMnetwork}, which can effectively deal with the vanishing gradient problem \cite{hochreiter1997lstm} as it allows for the gradient to flow unchanged. Additionally, since the weights are shared across all time steps, the total number of parameters to train is far less than the \ac{FC} structure. These features allows the algorithm to scale to optimal problems of long time horizons. Intuitively, one can also think of the use of \ac{LSTM} as modeling the time evolution of $V_{\tilde{x}}$, in contrast to the \ac{FC} structure, which acts independently at every time step.

\section{Simulation Results}\label{experiments}
We applied the Deep FBSDE controller to systems of pendulum, cartpole and quadcopter for the task of reaching a target final state. The trained networks are evaluated over 128 trials and the results are compared between the different network architectures for both the unconstrained and control constrained case. We use \ac{FC} and \ac{LSTM} to denote experiments with the network architectures in fig. \ref{fig:FCnetwork} and \ref{fig:LSTMnetwork} respectively. We use 2 layer \ac{FC} and \ac{LSTM} networks and tanh activation for all experiments, with $\Delta t=0.02\,s$. All experiments were conducted in TensorFlow \cite{tensorflow} on an Intel i7-4820k CPU Processor. A comparison of training time and trainable parameter number is shown in fig. \ref{fig:run_time}, where it is clear that the \ac{LSTM} network saves at least 20\% of training time and has much fewer parameters than the \ac{FC} network.

In all trajectory plots, the solid line represents the mean trajectory, and shaded region shows the 95\% confidence region. To differentiate between the 4 cases, we use \textcolor{blue}{blue for unconstrained \ac{FC}}, \textcolor{green}{green for unconstrained LSTM}, \textcolor{cyan}{cyan for constrained FC} and \textcolor{magenta}{magenta for constrained LSTM}.

\subsection{Pendulum}
The algorithm was applied to the pendulum system for the swing-up task with a time horizon of 1.5 seconds. The equation of motion for the pendulum is given by
\begin{align}
    ml^2\ddot{\theta}+mgl\sin{\theta}+b\dot{\theta}=u.
\end{align}
The initial pendulum angle is 0 $radian$, and the target pendulum angle and rate are $\pi$ $radians$ and 0 $rad/s$ respectively. A maximum torque constraint of $u^{max}=10\;Nm$ is used for the control constrained cases.

\begin{figure}
  \includegraphics[width=\linewidth]{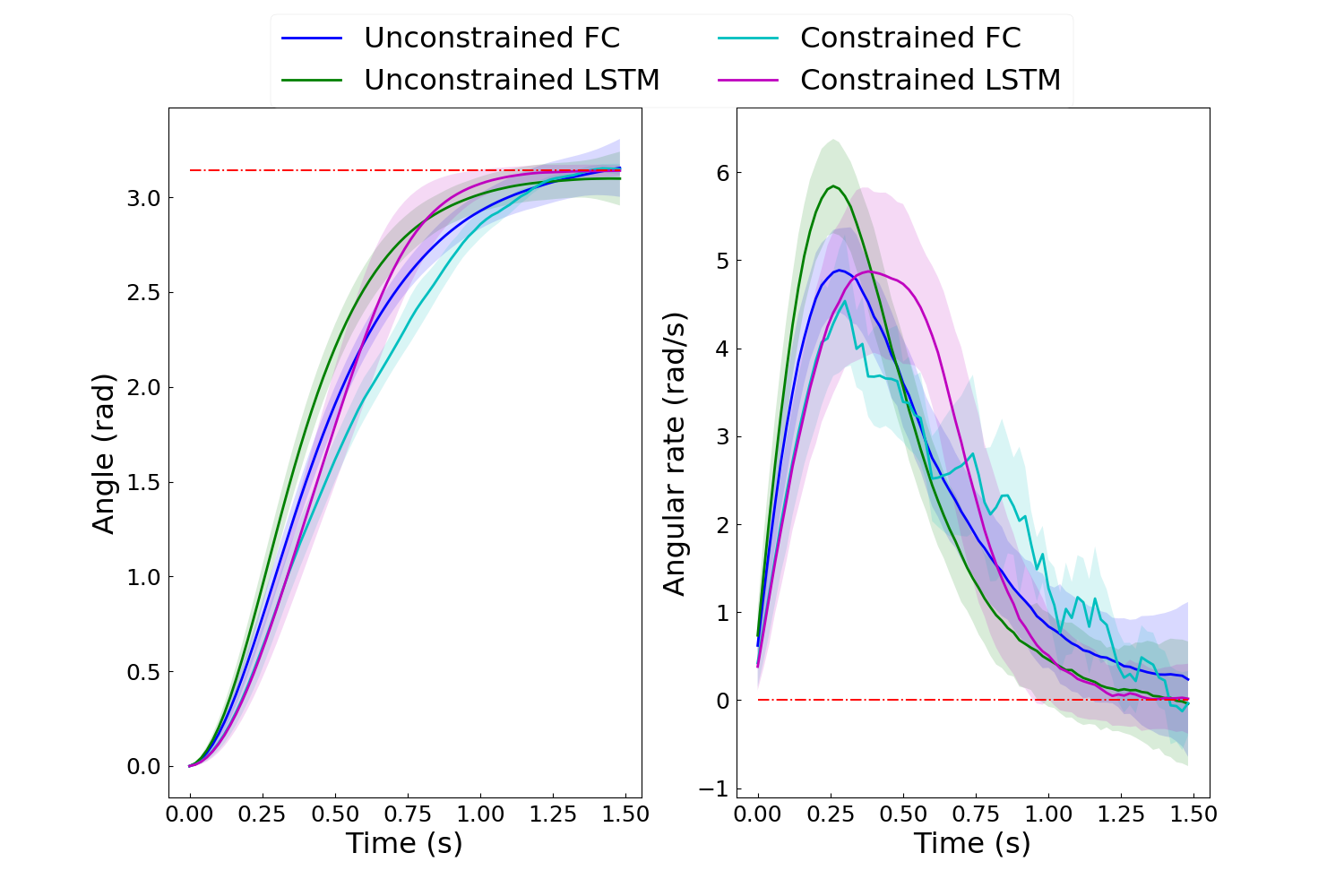}
  \caption{Pendulum states. \textit{Left}: Pendulum Angle; \textit{Right}: Pendulum Rate.}
  \label{fig:pendulumStates}
\end{figure}

\begin{figure}
  \includegraphics[width=\linewidth]{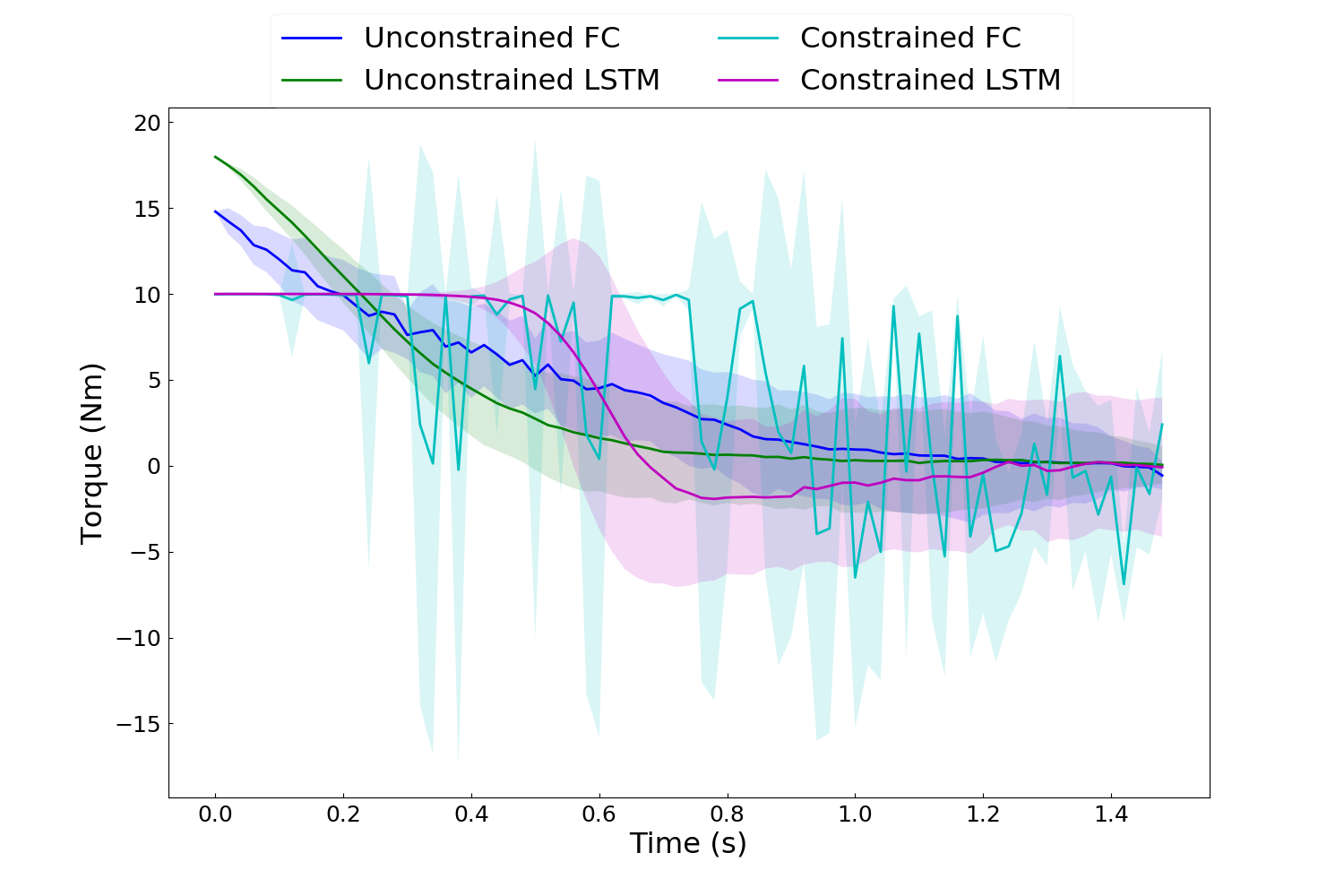}
  \caption{Pendulum controls.}
  \label{fig:pendulumControls}
\end{figure}

Fig. \ref{fig:pendulumStates} shows the state trajectories across the 4 case. It can be observed that the swing-up task is completed in all casess with low variance. However, the pole rate does not return to 0 for unconstrained \ac{FC}, as compared to unconstrained \ac{LSTM}. When the control is constrained, the pendulum angular rate becomes serrated for \ac{FC} while remaining smooth for \ac{LSTM}. This also more noticeable in the control torques (fig. \ref{fig:pendulumControls}). The control torques becomes very spiky for \ac{FC} due to the independent networks at each time step. On the other hand, the hidden temporal connection within \ac{LSTM} allows for smooth and optimally behaved control policy.


\subsection{Cart Pole}
The algorithm was applied to the cart-pole system for the swing-up task with a time horizon of 1.5 seconds. The equations of motion for the cart-pole are given by
\begin{align}
    2\ddot{x}+\ddot{\theta}\cos{\theta}-\dot{\theta}^2\sin{\theta}=u \\
    \ddot{x}\cos{\theta}+\ddot{\theta}+\sin{\theta}=0.
\end{align}
The initial pole angle is 0 $radian$, and the target pole angle is $\pi$ $radians$ with target pole and cart velocities of 0 $rad/s$ and 0 $m/s$ respectively. Note that despite the target of 0 $m$  for cart position, we do not penalize non-zero cart position in training. A maximum force constraint of 10 $N$ is used for the control constrained case.

\begin{figure}
  \includegraphics[width=\linewidth]{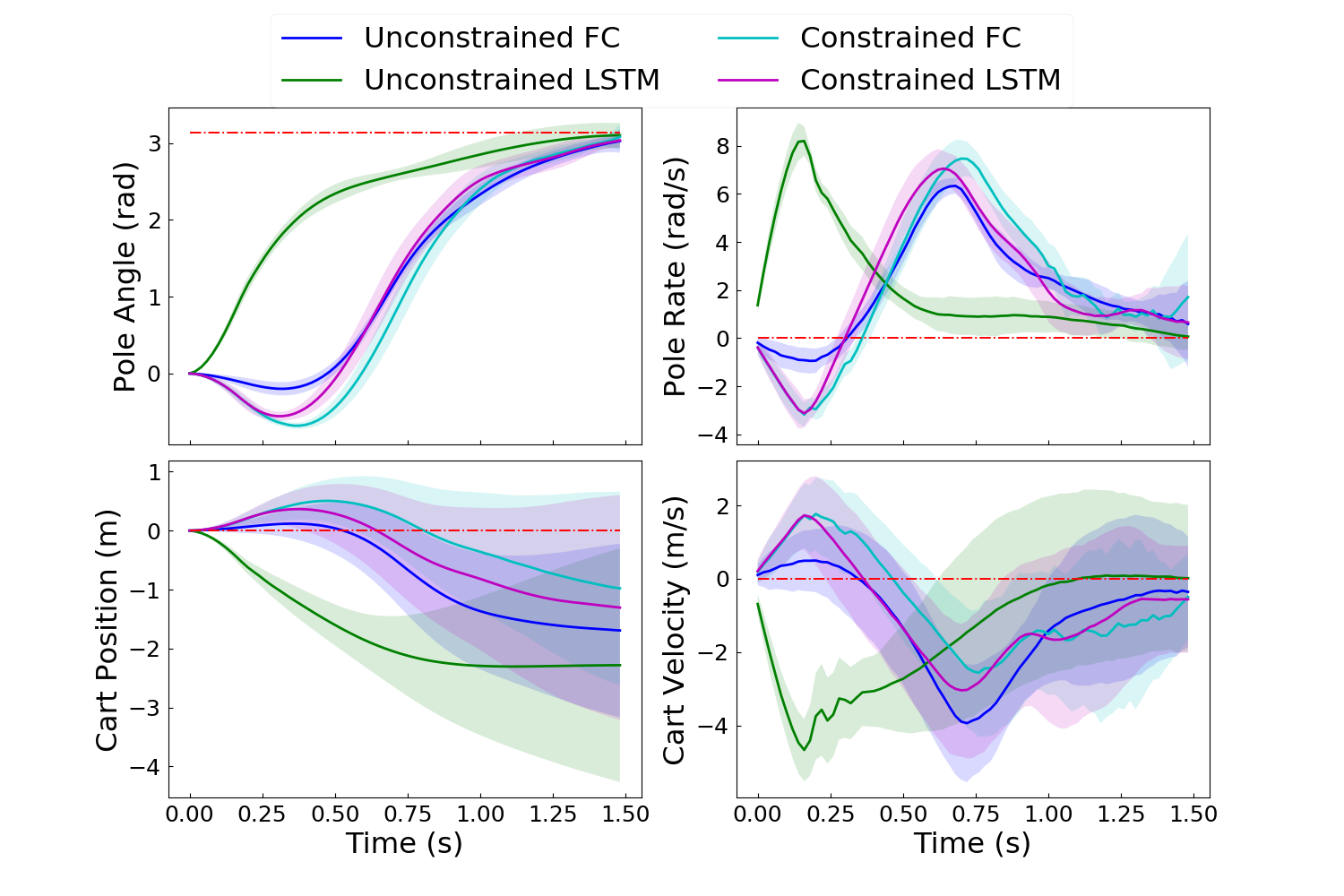}
  \caption{Cart Pole states. \textit{Top Left}: Pole Angle; \textit{Top Right}: Pole Rate; \textit{Bottom Left}: Cart Position; \textit{Bottom Right}: Cart Velocity.}
  \label{fig:CartPoleStates}
\end{figure}

The cart-pole states are shown in fig. \ref{fig:CartPoleStates}. Similar to the pendulum experiment, the swing-up task is completed with low variance acrossed all cases. Interestingly, when control is constrained, both \ac{FC} and \ac{LSTM} swing the pole in the direction opposite to target at first and utilize momentum to complete the task. Another interesting observation is that in the unconstrained case, the LSTM-policy is able to exploit long-term temporal connections to initially apply large controls to swing-up the pole and then focus on decelerating the pole for the rest of the time horizon, whereas the \ac{FC}-policy appears to be more myopic resulting in a delayed swing-up action. Similar to the pendulum experiment, under control constraint the \ac{FC}-policy results in sawtooth-like controls while the \ac{LSTM}-policy outputs smooth control trajectories.

\begin{figure}
  \includegraphics[width=\linewidth]{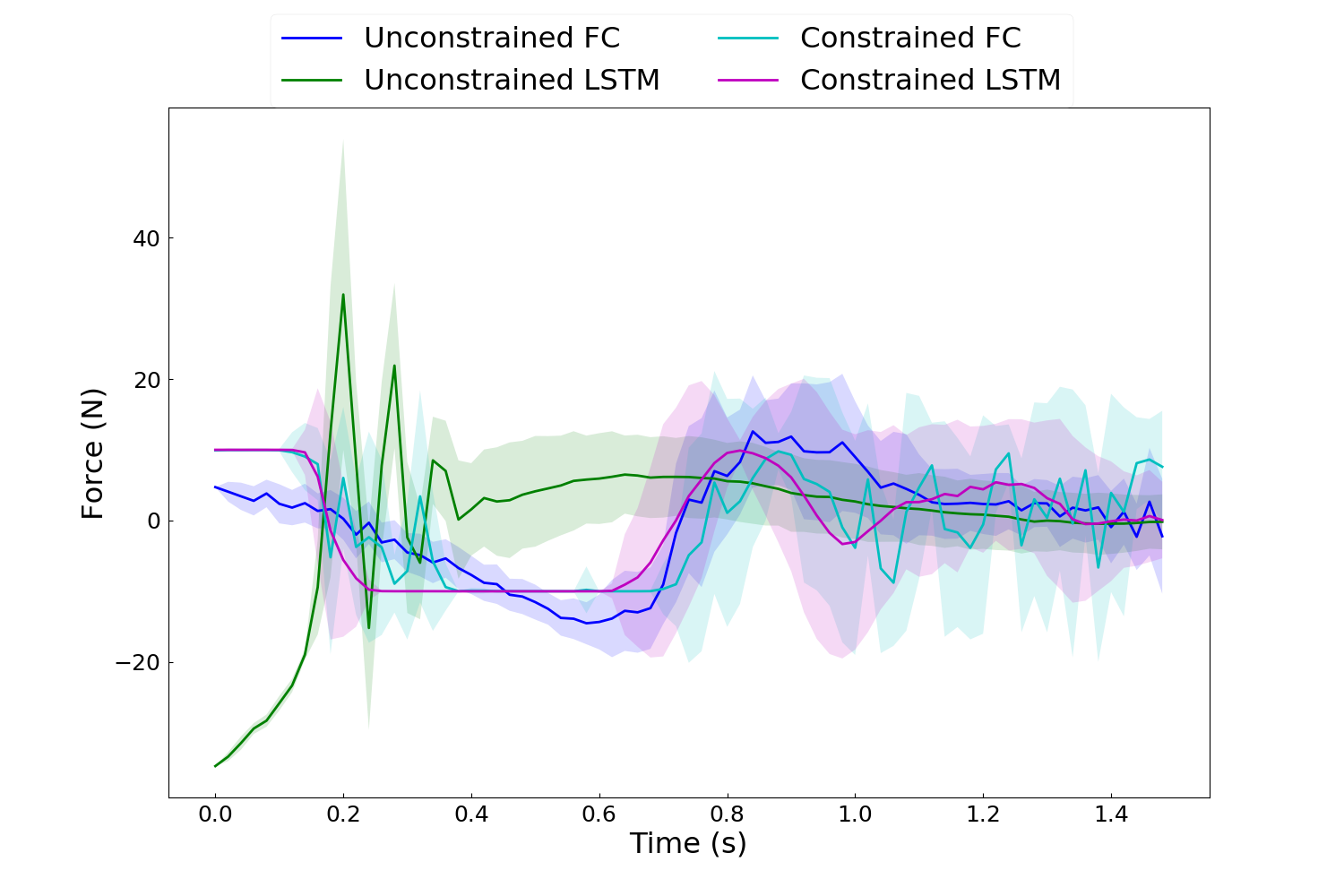}
  \caption{Cart Pole controls.}
  \label{fig:CartPoleControls}
\end{figure}

\subsection{Quadcopter}
The algorithm was applied to the quadcopter system for the task of flying from its initial position to a target final position with a time horizon of 2 seconds. The quadcopter dynamics used is described in detail by \citet{quad_dynamics}. The initial condition is 0 across all states, and the target is 1 $m$ upward, forward and to the right from the initial location with zero velocities and attitude. The controls are motor torques. A maximum torque constraint of 3 $Nm$ is imposed for the control constrained case. 

\begin{figure}
  \includegraphics[width=\linewidth]{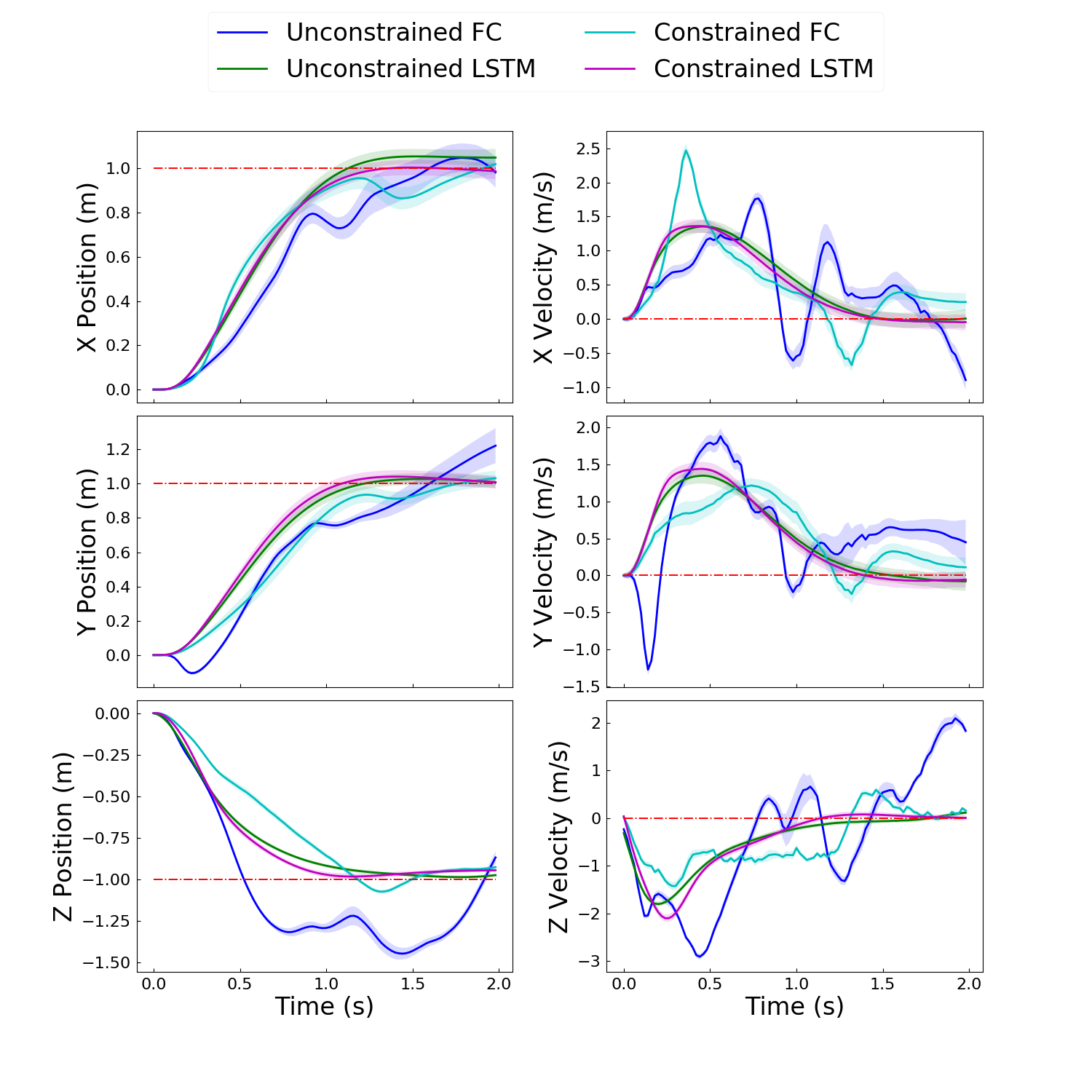}
  \caption{Quadcopter states. \textit{Top Left}: X Position; \textit{Top Right}: X Velocity; \textit{Middle Left}: Y Position; \textit{Middle Right}: Y Velocity; \textit{Bottom Left}: Z Position; \textit{Bottom Right}: Z Velocity.}
  \label{fig:QuadStates1}
\end{figure}

\begin{figure}
  \includegraphics[width=\linewidth]{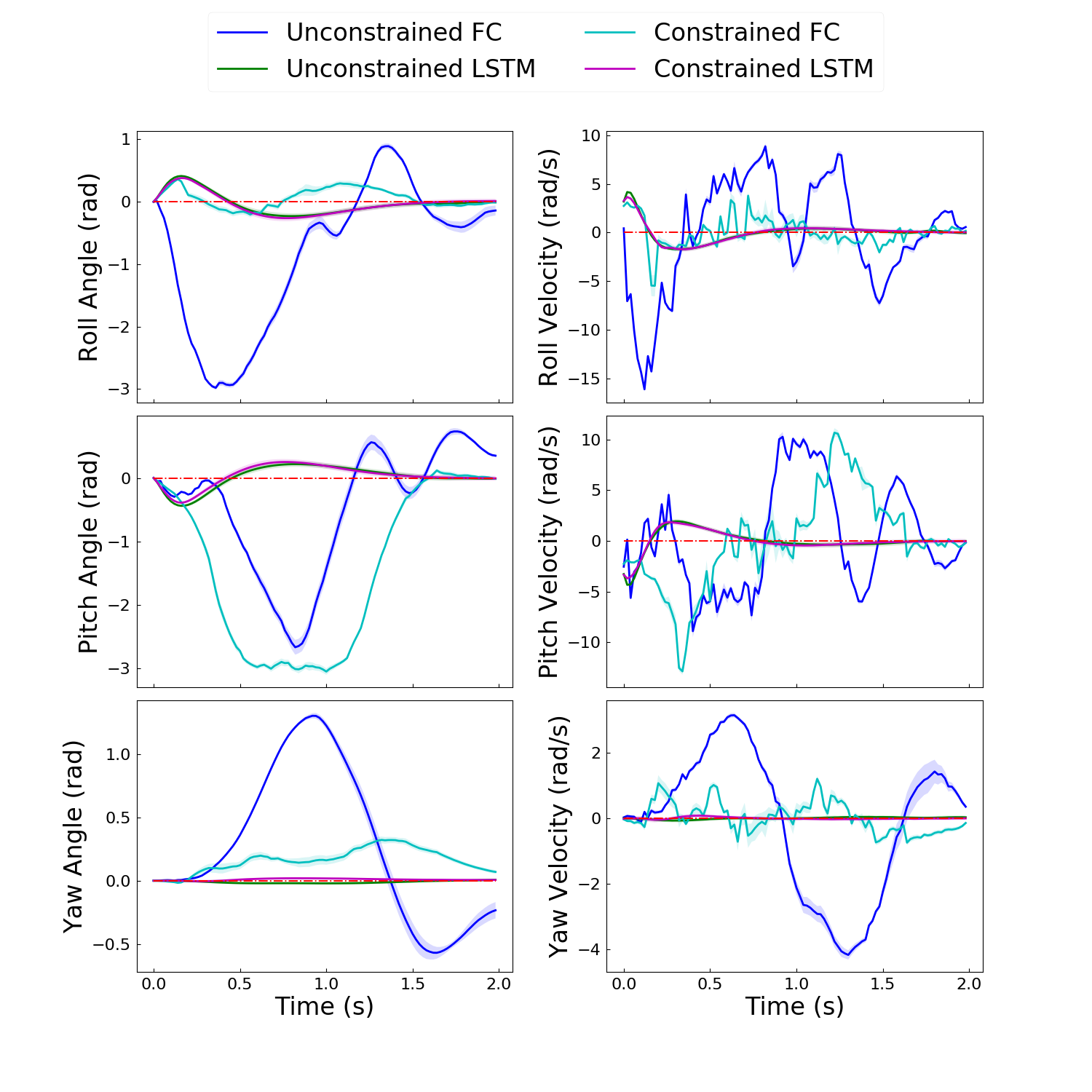}
  \caption{Quadcopter states. \textit{Top Left}: Roll Angle; \textit{Top Right}: Roll Velocity; \textit{Middle Left}: Pitch Angle; \textit{Middle Right}: Pitch Velocity; \textit{Bottom Left}: Yaw Angle; \textit{Bottom Right}: Yaw Velocity.}
  \label{fig:QuadStates2}
\end{figure}

\begin{figure}
  \includegraphics[width=\linewidth]{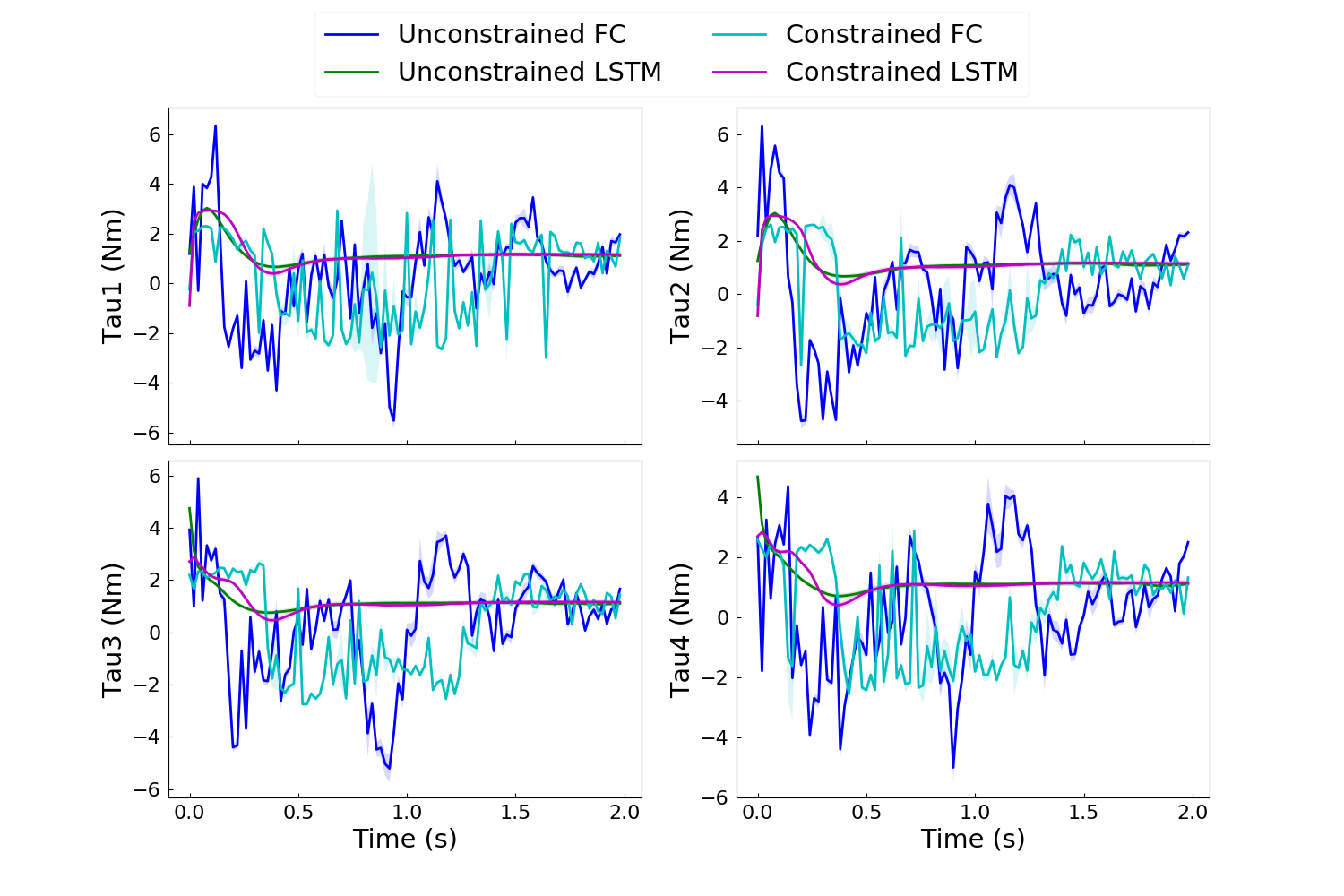}
  \caption{Quadcopter controls.}
  \label{fig:QuadControls}
\end{figure}

This task required $N=100$ individual \ac{FC} networks. After extensive experimentation, we conclude that tuning the \ac{FC}-based policy becomes significantly difficult and cumbersome as the time horizon of the task increases. On the other hand, tuning our proposed \ac{LSTM}-based policy was equivalent to that for the cart-pole and pendulum experiments. Moreover, the shared weights across all time steps results in faster build-times and run-times of the TensorFlow computational graph. As seen in the figures (\ref{fig:QuadStates1}-\ref{fig:QuadControls}) from our experiments, the performance of the \ac{LSTM}-based policies surpassed that of the \ac{FC}-based policies (especially for the attitude states) due to exploiting long term temporal dependence and ease of tuning.         


\section{Conclusions} \label{conclusion}
In this paper, we proposed the Deep FBSDE Control algorithm that utilizes fully  connected and recurrent layers based on the LSTM  network architecture. The proposed algorithm  solves finite time horizon \acl{SOC} problems for nonlinear systems with control-affine dynamics and constraints in the controls. 

  The architectures presented in this paper can be extended in many different ways, some of which include: 

\begin{itemize}
    \item \textbf{Risk-Sensitive and Min-Max \acl{SOC}:}  This type of \acl{SOC} problems result in the so-called Hamilton-Jacobi-Bellman-Isaacs PDE. The min-max formulations are typically used to model stochastic disturbances with unknown mean. Solving these \ac{SOC} problems will result in robust policies in robotics. 
    
    \item \textbf{\acl{SOC} of systems with generalized stochasticities:} For systems with L\'{e}vy and jump-diffusion noise, the resulting HJB equation is a partial-integro-differential equation. Stochastic models that include jump-diffusions could be used to model wind-gust or  ground forces in terrestrial vehicles. 
    
    \item \textbf{Non-affine control dynamics:} Very often in robotics dynamics are represented by function approximators such as \acp{DNN}
    or  \ac{GPs}. This choice  results in dynamics that are non-affine in controls. A potential new direction is to generalize the Deep FBSDE Control algorithm for such representations. 
\end{itemize}

\section*{Acknowledgments}
This research was supported by the Amazon Web Services Machine Learning Research Awards and the NSF CMMI award \#1662523.

\bibliographystyle{unsrtnat}
\bibliography{references,AdditionalReferences}

\end{document}